\pdfoutput=1

\documentclass[11pt]{article}

\usepackage{acl}

\usepackage{times}
\usepackage{latexsym}

\usepackage[T1]{fontenc}

\usepackage[utf8]{inputenc}

\usepackage{microtype}

\usepackage{graphicx}
\usepackage{float}
\usepackage[section]{algorithm}
\usepackage{algpseudocode}
\usepackage{pythonhighlight}
\usepackage{amssymb}
\usepackage{amsmath}
\usepackage{multirow}
\usepackage{booktabs}
\usepackage{comment}

\usepackage{amsmath}
\usepackage{oubraces}

\usepackage{enumitem}

\title{Prompting Large Pre-trained Vision-Language Models For Compositional Concept Learning}

\author{Guangyue Xu \\
  Michigan State University \\
  \\\And
  Parisa Kordjamshidi \\
  Michigan State University\\
  \\\And
  Joyce Chai\\
  University of Michigan\\
  }

\begin{document}
\maketitle
\begin{abstract}
This work explores the zero-shot compositional learning ability of large pre-trained vision-language models(VLMs)  within the prompt-based learning framework and propose a model (\textit{PromptCompVL}) to solve the compositonal zero-shot learning (CZSL) problem.
\textit{PromptCompVL} makes two design choices: first, it uses a soft-prompting instead of hard-prompting to inject learnable parameters to reprogram VLMs for compositional learning.
Second, to address the compositional challenge, it uses the soft-embedding layer to learn primitive concepts in different combinations. By combining both soft-embedding and soft-prompting, \textit{PromptCompVL} achieves state-of-the-art performance on the MIT-States dataset. Furthermore, our proposed model achieves consistent improvement compared to other CLIP-based methods which shows the effectiveness of the proposed prompting strategies for CZSL.
\end{abstract}

\section{Introduction}

\begin{figure}[ht]
 \begin{center} 
  \includegraphics[width=0.5\textwidth]{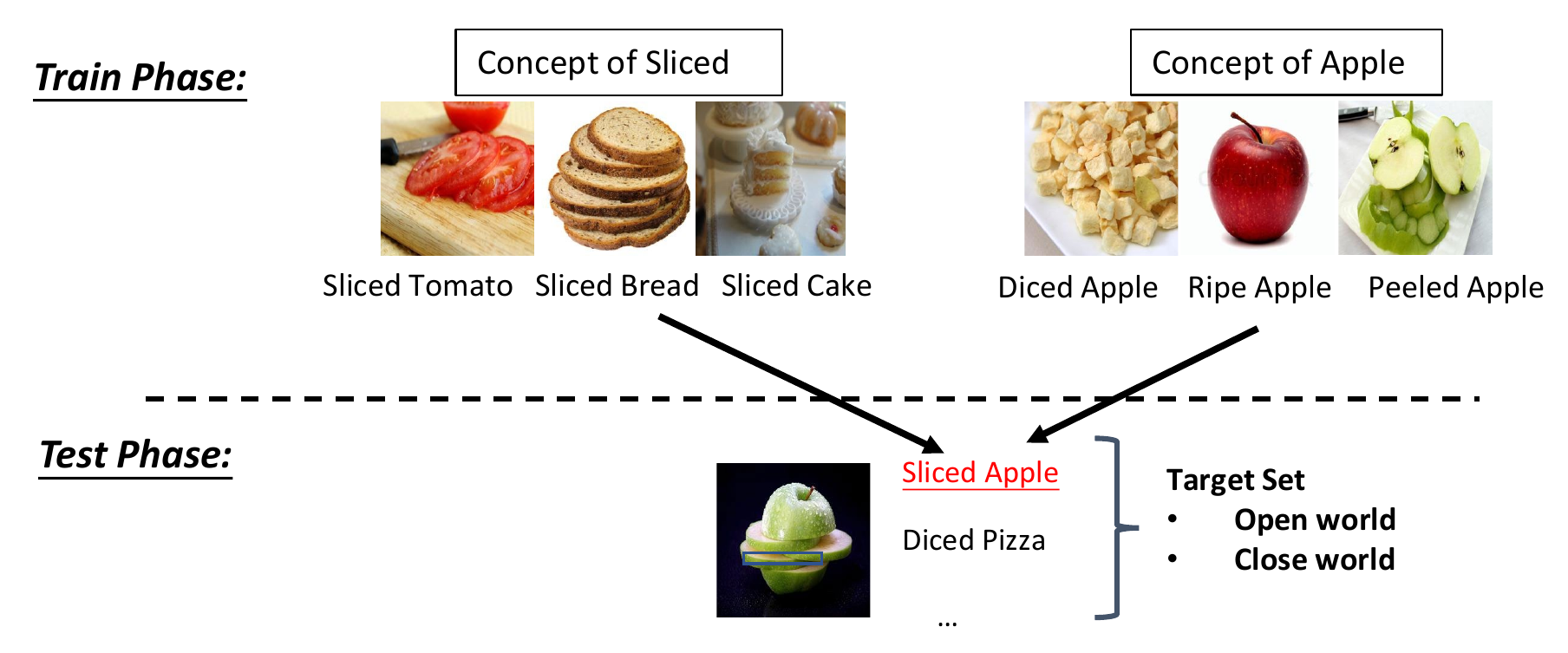}
  \caption{CZSL setting: given the primitive concepts of \textit{sliced} and \textit{apple}, our target is to recognize the compositional concept \textit{sliced apple}.} 
  \label{fig:example}
  \end{center}
\end{figure}

In this paper, 
we investigate a previously formulated compositional learning problem, compositional zero-shot  learning (CZSL), which requires the agent to recognize novel compositional attribute-object (attr-obj) pairs 
by composing previously learnt primitive concepts. For example, in Fig.~\ref{fig:example}, after learning the primitive concepts, \textit{sliced} and \textit{apple}, CZSL expects the agent to recognize \textit{sliced apple} which has not been observed during training time.

The main challenge of CZSL is the distribution-shift  between the training and test data which causes the learnt models overfit the seen compositions.
Previous works usually construct a shared embedding space and add different constraints to regularize the space for compositional concept learning~\cite{attrasopt,graph_comp,open_world_comp}.
In this work, we attempt to solve the CZSL problem from the lens of prompting large vision-language models. We propose a model, called \textit{PromptCompVL}, to explore the compositional learning ability of current VLMs. 

The core idea of \textit{PromptCompVL} is to inject learnable pieces, including the soft-prompting layer and the soft-embedding layer,
to CLIP~\cite{clip} for compositional learning. In particular, soft-prompting layer is used to  replace the CLIP's hard-prompting vectors in order to increase CLIP's capacity and reprogramming it for CZSL by adjusting the soft-prompting vectors. Moreover, soft-embedding layer is introduced to replace CLIP's original vocabulary embedding layer to address the compositional concept learning challenge. We use the soft-embedding layer to encode the primitive concepts and update the concept embedding through observing different combinations during training time.  The role of soft-embedding is similar to verbalizer in the general prompt-learning framework~\cite{liupengfei}. 

The advantages of this work can be summarized as follows:
1) Inherited from prompting methods, \textit{PromptCompVL} is a \textit{parameter-efficient} learning framework which can improve CZSL using VLMs without the overhead of fine-tuning the entire model.
2) Different from previous prompting architecture, \textit{PromptCompVL} introduces two learnable components, soft-embedding and soft-prompting simultaneously, to address CZSL problems. In particular,  it introduces the soft-embedding layer to address the compositional challenge and the soft-prompting layer to improve the VLMs' flexibly to fit CZSL tasks.
3) \textit{PromptCompVL} achieves SOTA result on MIT-States dataset and shows consistent improvements compared to other CLIP-based methods on both MIT-States and UT-Zappos datasets.



\section{Preliminaries}
\textbf{CLIP}~\cite{clip} is a powerful Vision-Language model which uses contrastive loss to learn a joint embedding space and align images and texts within the constructed space.  
CLIP consists of three components: 1) a text encoder to summarize text into a vector. It uses BERT~\cite{bert} as its text encoder, 2) an image encoder to transform image into a vector.  The image encoder can be ResNet~\cite{resnet} or ViT~\cite{vit}, 3) a loss function, which is contrastive loss to update the text and image encoders. In \textit{PromptCompVL}, we use and fix the pre-trained text and image encoders and solve CZSL by adding the learnable soft-prompting and soft-embedding layers detailed in Sec.~\ref{sec:arc}.

\noindent\textbf{Prompt Learning} is commonly used for transferring knowledge from pre-trained models to downstream tasks, especially in low-resource scenarios.
Prompt Learning can be generally categorized into: 
1) \textit{discrete/hard} prompting which requires carefully engineered prompts and verbalizers to map the vocabulary-space to the label-space for downstream tasks~\cite{gpt_understand_too}.
2) \textit{continuous/soft} prompting which injects learnable prompts into VLMs and reprogram the VLMs for downstream tasks. \textit{PromptCompVL} follows the soft-prompting line. Besides soft-prompting layer,  \textit{PromptCompVL} also adds soft-embedding layer to further improve CLIP's compositional learning ability.
Because CZSL has no training examples for novel attr-obj compositions, prompting large VLMs can help solve this zero-shot problem.


\section{Problem Formulation\label{problem_setting}}
Here we formally define the CZSL task.
Let $\mathbb{A}=\left\{a_{0}, a_{1}, \ldots, a_{n}\right\}$ be the attribute set and $\mathbb{O}=\left\{o_{0}, o_{1}, \ldots, o_{m}\right\}$ be the object set. The compositional label space $\mathbb{Y}$ is the Cartesian product of the attribute set and the object set, $\mathbb{Y}=\mathbb{A} \times \mathbb{O}$.
At training time, we are given seen examples  $\mathbb{S}_{\text {seen }}=\left\{\left(x_{1}, y_{1}\right), \ldots,\left(x_{n}, y_{n}\right)\right\}$, where $x_i$ is an image and $y_i=(a_i,o_i)$ is its label from seen pair set $\mathbb{Y}_{seen}  \subset \mathbb{Y}$. 
The goal of CZSL is to learn a function $f$ to assign an image a compositional label from the target set $\mathbb{Y}_{target} \subseteq \mathbb{Y}$. 
Based on different target set settings, CZSL can be categorized into: 1) Standard CZSL, where $\mathbb{Y}_{target} = \mathbb{Y}_{unseen}$ and $\mathbb{Y}_{seen} \cap \mathbb{Y}_{unseen}=\varnothing$, the target set only consists of unseen pairs introduced in \cite{attrasopt}; 2) Generalized CZSL, where $\mathbb{Y}_{target} = \mathbb{Y}_{seen} \cup \mathbb{Y}_{unseen}$, the target set consists of both seen and unseen pairs introduced in \cite{modu_net}; 3) Open-world CZSL where $\mathbb{Y}_{target} = \mathbb{Y}$ where target set is all attr-obj combinations and is the most challenging case introduced in \cite{open_world_comp}. Following \cite{modu_net}'s pair split, we evaluate \textit{PromptCompVL} using generalized CZSL as the close world and open-world CZSL as the open world detailed in Appendix \ref{czsl_setting}. 



\begin{figure*}[t]
\includegraphics[width=\textwidth]{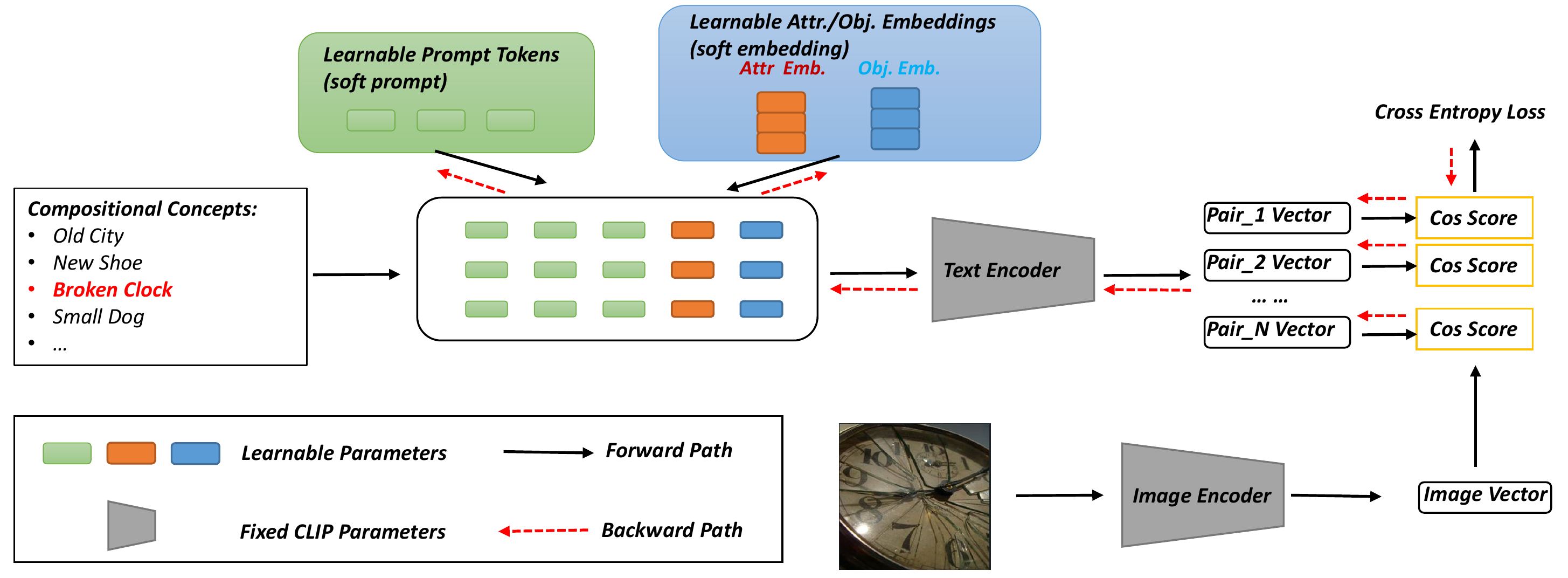}
\centering
\caption{\textit{PromptCompVL} Architecture. It consists of four components: image encoder, text encoder, soft-prompting layer and soft-embedding layer. The Soft-prompting and soft-embedding layers are learnable during training.}
\label{fig:arch}
\end{figure*}

\begin{figure}[ht]
\begin{center} 
\includegraphics[width=0.9\linewidth]{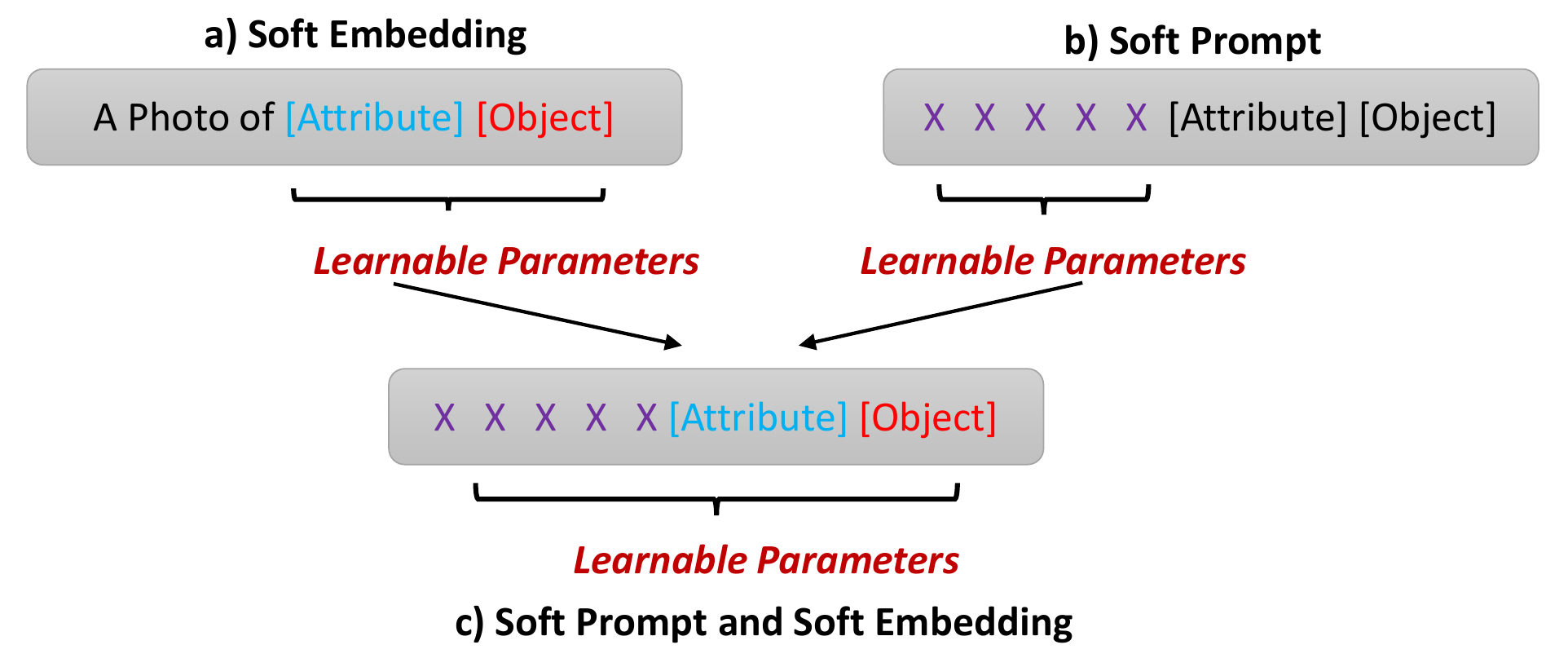}
\end{center}
\caption{Different prompting strategies. \textit{PromptCompVL} combines both soft-prompting and soft-embedding.}
\label{promptType}
\end{figure}

\section{PromptCompVL}

\subsection{Architecture\label{sec:arc}}
The architecture of \textit{PromptCompVL} is shown in Fig.~\ref{fig:arch}. It has four components: text encoder, image encoder, soft-embedding layer and soft-prompting layer. As Fig.~\ref{promptType} shows, different from previous prompting strategies~\cite{coop,csp}, we set both soft-prompting and soft-embedding as learnable parameters and we use these two soft layers to construct text input as Eq.~\ref{eq:prompt}. 

\begin{equation}
\setlength\abovedisplayskip{0pt}
[\underbrace{SOS, \overbrace{v_1, v_2, ...,v_k}^{\textbf{soft-prompt}}, \overbrace{attr, obj,}^{\textbf{soft-embedding}} EOS}_{\textbf{Text Context}}]
\label{eq:prompt}
\end{equation}

\begin{itemize}
    \item \textit{Soft-embedding layer}. We update the embedding for each primitive concept under different combinations during training time. And this improve the model's compositional ability. The soft-embedding layer is with size $R^{(|a|+|o|)*d}$, where $|a|$ is the attribute number, $|o|$ is the object number, $d$ is the soft-embedding dimension.  
    \item \textit{Soft-prompting layer}. In this setting, by adding the soft-prompting parameters, we reprogram CLIP  for different compositional learning datasets. 
    We have $k$ prompt vectors, each with dimension $d$, i.e.,  $v_i \in R^d, i \in {1,2,...k}$.  
\end{itemize}





\subsection{Pipeline}
In this section, we will discuss the details of \textit{PromptCompVL}'s components and the whole pipeline is outlined in Appendix~\ref{app:pipeline}. 


\noindent\textbf{Text Encoder.} 
Given the attr-obj label,
1) instead of using CLIP's hard prompts, \textit{"a photo of"}, we add the learnable soft-prompting vectors $[v_1, v_2, ..., v_k]$ before the attr-obj label, 2) we encodes attr-obj label using the learnable soft-embedding layer instead of the CLIP's fixed embedding layer, 3) after the above replacements, we extract and normalize \textit{EOS} vector $t_i$ after self-attention mechanism as the text representation using  Eq.~\ref{eq:encoder}.



\noindent\textbf{Image Encoder.} 
We choose ViT 
as our backbone image encoder. We first rescale the image to size $224 \times 224$. Then we use ViT to extract the image vector. Specifically, \textit{ViT-L/14} is used as \textit{PromptCompVL}'s image encoder. The extracted image vector $x_i$ needs to be normalized according to Eq.~\ref{eq:encoder}.

\begin{table*}[ht]
\small
\centering
\resizebox{\textwidth}{!}{
\begin{tabular}{ccccccccccccccccc} 
\hline
& \multicolumn{8}{c}{\textbf{Close World}} & \multicolumn{8}{c}{\textbf{Open World}}\\
\cmidrule(r){2-9}\cmidrule(r){10-17}
& \multicolumn{4}{c}{\textbf{Mit-States}} & \multicolumn{4}{c}{\textbf{UT-Zappos}}  & \multicolumn{4}{c}{\textbf{Mit-States}} & \multicolumn{4}{c}{\textbf{UT-Zappos}}\\
\cmidrule(r){2-5}\cmidrule(r){6-9}\cmidrule(r){10-13}\cmidrule(r){14-17}
\textbf{Method} & S & U & H & AUC & S & U & H & AUC & S & U & H & AUC & S & U & H & AUC\\
\hline
\textbf{AoP}~\cite{attrasopt} & 14.3 & 17.4 & 9.9 & 1.6 & 59.8 & 54.2 & 40.8 & 25.9 & 16.6 & 5.7 & 4.7 & 0.7 & 50.9 & 34.2 & 29.4 & 13.7\\
\textbf{LE+}~\cite{redwine} & 15.0 & 20.1 & 10.7 & 2.0 & 53.0 & 61.9 & 41.0 & 25.7 & 14.2 & 2.5 & 2.7 & 0.3 & 60.4 & 36.5 & 30.5 & 16.3\\
\textbf{TMN}~\cite{modu_net} & 20.2 & 20.1 & 13.0 & 2.9 & 58.7 & 60.0 & 45.0 & 29.3 & 12.6 & 0.9 & 1.2 & 0.1 & 55.9 & 18.1 & 21.7 & 8.4\\
\textbf{SymNet}~\cite{sym_net} & 24.2 & 25.2 & 16.1 & 3.0 & 49.8 & 57.4 & 40.4 & 23.4 & 21.4 & 7.0 & 5.8 & 0.8 & 53.3 & 44.6 & 34.5 & 18.5\\
\textbf{CompCos}~\cite{open_world_comp} & 25.3 & 24.6 & 16.4 & 4.5 & 59.8 & 62.5 & 43.1 & 28.7 & 25.4 & 10.0 & 8.9 & 1.6 & 59.3 & 46.8 & 36.9 & 21.3\\
\textbf{CGE}~\cite{graph_comp} & 32.8 & 28.0 & 21.4 & 6.5 & \textbf{64.5} & \textbf{71.5} & \textbf{60.5} & \textbf{33.5} & 32.4 & 5.1 & 6.0 & 1.0 & 61.7 & \textbf{47.7} & \textbf{39.0} & \textbf{23.1} \\
\hline
\textbf{CLIP}~\cite{clip} & 30.2 & 40.0 & 26.1 & 11.0 & 15.8 & 49.1 & 15.6 & 5.0 & 30.1 & 14.3 & 12.8 & 3.0 & 15.7 & 20.6 & 11.2 & 2.2\\
\textbf{COOP}~\cite{coop} & 36.7 & 49.2 & 31.6 & 15.1 & 62.9 & 62.3 & 45.5 & 31.3 & 36.8 & \textbf{16.5} & 16.1 & 4.7 & 61.8 & 39.3 & 35.6 & 19.5\\ 
\textbf{CSP}~\cite{csp} & 48.2 & 45.64 & 34.4 & 17.9 & 63.8 & 64.0 & 45.2 & 31.9 & 46.3 & 15.7 & 17.4 & 5.7 & 64.1 & 44.1 & 38.9 & 22.7\\
\hline
\textbf{Ours} & \textbf{48.5} & \textbf{47.2} & \textbf{35.3} & \textbf{18.3} & 64.4 & 64.0 & 46.12 & 32.2 & \textbf{48.5} & 16.0 & \textbf{17.7} & \textbf{6.1} & \textbf{64.6} & 44.0 & 37.1 & 21.6\\
\hline
\end{tabular}}
\caption{Close and Open World results on UT-Zappos and Mit-States datasets. We report best seen accuracy S, best unseen accuracy U, best harmonic mean(HM) and area under the curve(AUC) for comparison.}
\label{tab:result}
\end{table*}

\noindent\textbf{Soft-Prompting and Soft-Embedding.} These are the components that we need to learn in the \textit{PromptCompVL} model.
We use these two components to construct the text context for each attr-obj pairs as CLIP's input as shown in Eq.~\ref{eq:prompt}. We use $\theta$ and $\phi$ to denote the learnable parameters for soft-prompting and soft-embedding respectively.

\begin{equation}
\setlength\abovedisplayskip{0pt}
\boldsymbol{x}_{i}=\frac{VisEnc(x_i)}{\left\|VisEnc(x_i)\right\|}
\quad 
\boldsymbol{t}_{i}=\frac{TxtEnc_{\theta,\phi}\left(t_{a_i,o_i}\right)}{\left\|TxtEnc_{\theta,\phi}\left(t_{a_i, o_i}\right)\right\|}
\label{eq:encoder}
\end{equation}

\subsection{Training}
After obtaining attr-obj and image vectors following the aforementioned steps, we can calculate the class probability using the cosine similarity as  Eq.~\ref{eq:sim}. Finally,  Cross-Entropy loss is used to update  PromptCompVL's soft-prompting parameters $\theta$ and soft-embedding parameters $\phi$ using the training dataset.


\begin{equation}
\setlength\abovedisplayskip{0pt}
p(y \mid x)=\frac{\exp \left(\operatorname{sim}\left(x, t_i\right) / \tau\right)}{\sum_{i=1}^{K} \exp \left(\operatorname{sim}\left(x, t_i\right) / \tau\right).}
\label{eq:sim}
\end{equation}

\noindent where $\tau$ is a temperature hyper-parameter, $sim$ denotes cosine similarity and $K$ is the number of attr-obj pairs in the training set.

\subsection{Inference}
Given an image, for each attr-obj pair in $\mathbb{Y}_{target}$, we construct the text input using the learnt soft-embedding and soft-prompting as the format of Eq.~\ref{eq:prompt}. After going through CLIP's text and image encoders, we use cosine similarity to select the most relevant attr-obj pair as the compositional label of the given image as follows,

\begin{equation}
\setlength\abovedisplayskip{0pt}
\hat{y}=\underset{t_i \in \mathbb{Y}_{\text {target}}}{\arg \max }\ sim \left(t_i,x\right).
\label{eq:infer}
\end{equation}

\section{Experiments}

\subsection{DataSets}

MIT-States~\cite{mitstates} and UT-Zappos~\cite{zappos} are commonly used as the benchmarks for CZSL. In our experiments, we follow \cite{modu_net}'s pair split shown in Tab.~\ref{data_split}. 


\subsection{Results}

We compare \textit{PromptCompVL} with two types of baselines: 1) non-CLIP methods (top six models) and 2) CLIP-based methods (the bottom three models) as shown in Table~\ref{tab:result}. 

In the open setting, due to distribution-shift problem, we need to introduce the feasibility score to adjust the unseen pairs' logits as used in ~\cite{open_world_comp,csp} and the adapted feasibility score is detailed in the Appendix~\ref{feas_score}. 

For fair comparison, the context length is set to $8$, the prompt length $k$ is set to $3$ which is CLIP's hard-prompting's length, both soft-embedding and soft-prompting's dimension $d$ is set to $768$ which is consistent with \textit{ViT-L/14}'s model setting.


\noindent\textbf{Results on MIT-States}. 
Table~\ref{tab:result} shows that  
\textit{PromptCompVL} achieves the new SoTA results on 
MIT-States on both close and open settings (except unseen accuracy in the open setting) compared with both CLIP and non-CLIP baselines.
The CLIP-based models have consistently better performance compared to the non-CLIP methods. Moreover, CLIP-prompting methods, including COOP, CSP and ours, further boost the performance compared to the vanilla CLIP model, which demonstrates the effectiveness of prompt learning in CZSL.


\noindent\textbf{Results on UT-Zappos}. Basically on UT-Zappos, \textit{PromptCompVL} and other CLIP-based approaches can't achieve SOTA performance compared with CGE.  It is likely because  UT-Zappos is a domain-specific dataset which consists of shoes and the materials. CLIP doesn't see many images from this domain during training time.


\begin{table}[]
\centering
\small
\setlength\tabcolsep{3.5pt}
\begin{tabular}{*{3}c}
\toprule
&  MIT-States & UT-Zappos\\
\hline
\# $Attr.$ & $115$ & $16$\\
\# $Obj.$ & $245$ & $12$\\
\# $Attr. \times Obj.$ & $28175$ & $192$\\
\hline
\# Train Pair & $1262$ & $83$\\
\# Train Img. & $30338$ & $22998$\\
\hline
\# Val. Seen Pair & $300$ & $15$ \\
\# Val. Unseen Pair & $300$ & $15$ \\
\# Val. Img. & $10420$ & $3214$\\
\hline
\# Test Seen Pair & $400$  & $18$\\
\# Test Unseen Pair & $400$  & $18$\\
\# Test Img. & $19191$ & $2914$\\
\hline
\end{tabular}
\caption{Dataset Split Statistics.}
\label{data_split}
\end{table}


\noindent\textbf{Comparing \textit{PromptCompVL} with other CLIP-based method}. Besides the absolute SOTA improvement on MIT-States, another interesting part is the \textit{PromptCompVL} gets consistent improvement on MIT-States in both settings and on UT-Zappos in close setting compared with other CLIP-based methods. This empirically shows the effectiveness of introducing both soft-embedding and soft-prompting in CZSL. Comparing with CSP~\cite{csp}, we only introduce additional $3$ learnable prompt vectors and obtain satisfactory improvements on MIT-States.
This shows the importance of soft-prompting. It reprograms CLIP for CZSL. For soft-embedding, we learn the primitive concept embedding instead of using fixed CLIP's embedding which is better for compositional learning compared with COOP~\cite{coop}.

\section{Conclusion}
In this paper, we propose \textit{PromptCombVL}, a new CLIP-based prompting framework,  to address the CZSL problem. Our initial results have shown that \textit{PromptCombVL} performs better for a wider domain such as MIT-States, however less effective for a more specific domain such as UT-Zappos. These results demonstrate potential advantages and limitations in applying CLIP-based prompting approaches to compositional concept learning in the future.   

\bibliography{anthology,custom}
\bibliographystyle{acl_natbib}

\appendix

\section{CZSL Settings\label{czsl_setting}}

\begin{figure}[ht]
 \begin{center} 
  \includegraphics[width=0.9\linewidth]{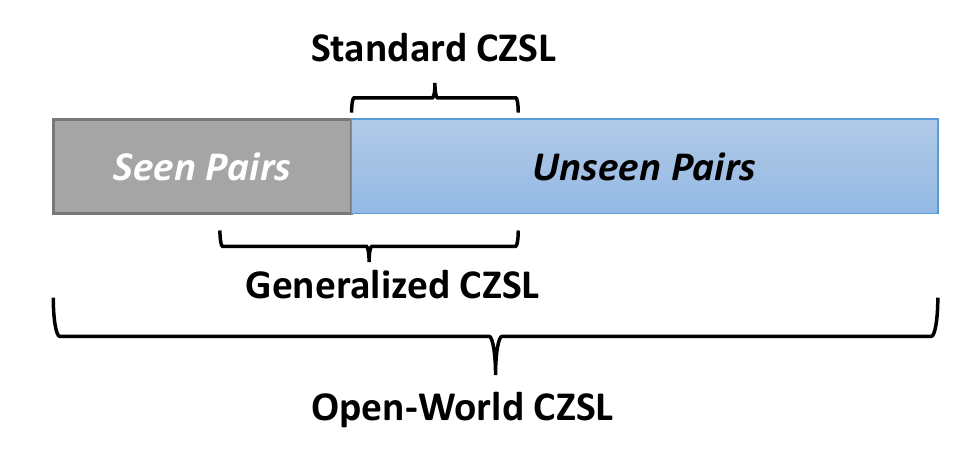}
  \caption{Illustration of different CZSL settings.}
  \label{fig:czsl_setting}
  \end{center}
\end{figure}

Fig.~\ref{fig:czsl_setting} shows the different target dataset choice in test phase. In our experiment, we select generalized CZSL as our close world setting and open-world CZSL as our open world setting used in recent works~\cite{csp,open_world_comp,graph_comp}.

\section{Pipeline\label{app:pipeline}}

\begin{algorithm}[H]
\begin{algorithmic}[1]
\State Initialize PromptCompVL using the pre-trained CLIP's text and image encoders.
\State Construct text input for attr-obj labels using Eq.~\ref{eq:prompt}.
\State Extract and normalize image/text vectors using CLIP's image/text encoder using Eq.~\ref{eq:encoder}.
\State Calculate the class probability as  Eq.~\ref{eq:sim} using the cosine similarity and update PromptCompVL's soft-prompting layer $\theta$ and soft-embedding layer $\phi$  using Cross-Entropy loss.
\end{algorithmic}
\caption{\textit{PromptComptVL}}
\label{alg}
\end{algorithm}

\section{Feasibility Scores\label{feas_score}}

\begin{table}[hb]
\centering
\begin{tabular}{*{2}c}
\toprule
Dataset &  Feasibility Score\\
\hline
MIT-States & 0.40691\\
UT-Zappos & 0.5299\\
C-GQA & \\
\hline
\end{tabular}
\caption{\textit{PromptCompVL}'s Feasibility score for Mit-States and UT-Zappos which is tuned using the validation set.}
\label{table:data_split}
\end{table}

\end{document}